\pdfoutput=1

\documentclass[11pt]{article}

\usepackage[]{acl}

\usepackage{times}
\usepackage{latexsym}

\usepackage[T1]{fontenc}

\usepackage[utf8]{inputenc}

\usepackage{microtype}

\usepackage{inconsolata}

\usepackage{multirow}

\usepackage[whole]{bxcjkjatype}
\usepackage{here}
\usepackage{amsmath}
\usepackage{amssymb}

\usepackage{booktabs}

\usepackage{graphicx}
\usepackage{tabularx}
\usepackage{soul}

\DeclareMathOperator*{\argmax}{arg\,max}

\DeclareMathOperator*{\NLL}{\mathrm{NLL}}

\title{Japanese-English Sentence Translation Exercises Dataset \\ for Automatic Grading}

\author{%
  Naoki Miura${}^{1,2}$　Hiroaki Funayama${}^{1,2}$　Seiya Kikuchi${}^{*,1,2}$\\{\bf Yuichiroh Matsubayashi${}^{1,2}$
　Yuya Iwase${}^{1,2}$　Kentaro Inui${}^{3,1,2}$}\\
${}^{1}$Tohoku University
　${}^{2}$RIKEN
　${}^{3}$MBZUAI \\ \texttt{\{miura.naoki.p6, h.funa, yuya.iwase.t8\}@dc.tohoku.ac.jp}\\
　\texttt{y.m@tohoku.ac.jp}
　\texttt{kentaro.inui@mbzuai.ac.ae}}

\begin{document}
\maketitle
\def\thefootnote{*}
\footnotetext{Work done while at RIKEN and Tohoku University. Currently belongs to NTT DATA INTELLILINK Corporation.}\def\thefootnote{\arabic{footnote}}

\begin{abstract}
This paper proposes the task of automatic assessment of Sentence Translation Exercises (STEs), that have been used in the early stage of L2 language learning.
We formalize the task as grading student responses for each rubric criterion pre-specified by the educators.
We then create a dataset for STE between Japanese and English including 21 questions, along with a total of $3,498$ student responses ($167$ on average).
The answer responses were collected from students and crowd workers.
Using this dataset, we demonstrate the performance of baselines including finetuned BERT and GPT models with few-shot in-context learning. 
Experimental results show that the baseline model with finetuned BERT was able to classify correct responses with approximately 90\% in $F_1$, but only less than 80\% for incorrect responses. 
Furthermore, the GPT models with few-shot learning show poorer results than finetuned BERT, indicating that our newly proposed task presents a challenging issue, even for the state-of-the-art large language models.
\end{abstract}

\section{Introduction}
\begin{figure}[t]
  \centering
  \includegraphics[width=75mm]{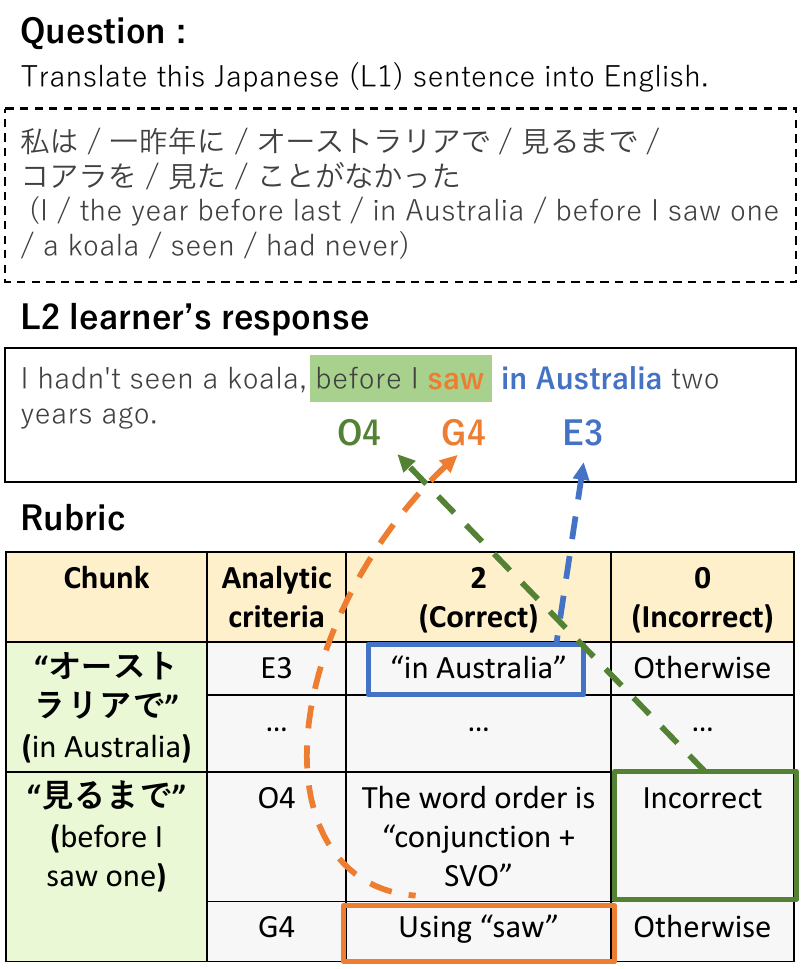}
  \caption{Example of sentence translation exercise. We excerpted the analytic criteria ``E3,'' ``O4,'' and ``G4'' from Q11 in our dataset. The correct answer is ``I had never seen a koala before I saw one in Australia two years ago.'' ``Chunk'' denotes a Japanese phrasal unit. ``E,'' ``O,'' and ``G'' are categories of each analytic criterion, which stand for ``expression,'' ``word order,'' and ``grammar,'' respectively.}
  \label{fig:sentence_translation_example}
\end{figure}

Sentence translation exercises (STEs) are often utilized as educational tools in the early stages of L2 language learning, particularly between language pairs that are linguistically distant from each other~\cite{oro21266,Butzkamm_bilingual_reform}.
Figure~\ref{fig:sentence_translation_example} shows an example of STE. Here, a learner translates a short sentence in their native language (L1) into the language they are learning (L2), and these translations are graded following analytic criteria within the grading rubric such as E3 and G4, which correspond to specific grammar items or expressions. 

This format facilitates the recognition of similarities and differences between the native language and the target language, which is especially effective in helping learners acquire basic grammar and expressions in the early stages of their language learning, thus enhancing their understanding of the desired modes of expression~\cite{oro21266}.
The questions in these exercises are brief and repeatable tests that efficiently help learners practice specific grammatical items, basic vocabulary, and idioms at a certain proficiency level and learn the nuances of expression between L1 and L2. 
Teachers can also use these exercises as assessment tools to evaluate whether learners have mastered specific grammar items or a vocabulary level.

However, because the responses to these exercises are descriptive, they pose a significant burden on educators in the form of manual grading and feedback.
Such a limitation restricts the frequency of these exercises despite the importance of repetitive training in language acquisition ~\cite{LarsenFreeman-2012}.
Therefore, automating the correction and feedback for translation exercises has the potential to significantly transform the educational environment in language learning.

Therefore, we aim to automate the grading of L1-to-L2 STEs. 
Tasks that are closely associated with this challenge include Grammatical Error Correction (GEC), which evaluates the grammatical correctness of written sentences, and machine translation.
STEs, however, are substantially different from these tasks in that they are usually operationalized with explicit learning objectives and closely reflect educators' intentions (\S\ref{ssec:sentence_tranlation_exercises}).
STEs not only clarify the learning objectives of a particular question but also allow for a more detailed learning analysis based on the performance of each evaluation item. 
The motivation for incorporating educators' intentions is also supported by studies that have found that the sole use of the GEC system does not elicit effective learner engagement~\cite{Koltovskaia2020-je,Ranalli2021-sl}.

To achieve our goal, we perform three tasks: (1) question formulation, (2) dataset creation, and (3) evaluation of baseline systems for our task.
To the best of our knowledge, this is the first attempt at an automated STE grading for educational purposes. 
Therefore, we first formulate the question. 
An important aspect of this formulation is to ensure that the established framework reflects the educators' evaluation criteria. 
Consequently, we formulate our task as a classification of scores on each evaluation item according to the predefined rubrics.
We then develop the dataset for this task. 
The questions and the rubric were created by English education experts, and answer scripts were collected from secondary education classrooms and through crowdsourcing.
Finally, we evaluate the performance of the conventional automated scoring model typically used for short answer scoring (SAS), as well as the latest generative language models with few-shot learning.

Experimental results showed that the baseline model using finetuned BERT successfully classified approximately 90\% of correct responses in $F_1$, but only less than 80\% of incorrect responses. 
Furthermore, GPT models with few-shot learning showed poorer results than the BERT model, indicating that even with a state-of-the-art LLM, our proposed new task remains difficult and challenging.
Error analysis of the few-shot models revealed their lack of comprehension regarding the grading task.

The contributions of this study are the following:
\begin{itemize}
    \item We formulate the automated grading of sentence translation exercises as a new task, referencing the actual operation of STEs in educational settings.
    \item We construct a dataset for the automated STE grading in accordance with this task design, which includes a total of 21 questions and $3,498$ responses, and demonstrate the feasibility of the task.
    \item We establish baseline performances for the task, showing potential for advancement.
\end{itemize}

\section{Automatic scoring of sentence translation exercises}
\label{sec:task}
\subsection{Sentence translation exercises}
\label{ssec:sentence_tranlation_exercises}
Sentence translation exercises (STEs) are a language learning tool where a learner translates a sentence in L1 into a target L2.
Studies have shown that the use of L1 in L2 education promots an understanding of differences and similarities between the two languages ~\cite{Butzkamm_bilingual_reform,oro21266}, reduces incomprehension, and enhances learning focus \cite{scott_whats_the_problem}.
Language translation has also been effective in improving students' four skills (speaking, writing, reading, listening) and promoting learning and communication skills \cite{yasar2022translanguaging}.
Because of these benefits, STEs are widely used in educational settings, particularly among beginners in language learning.

Figure \ref{fig:sentence_translation_example} shows an overview of an STE.
A learner's translated response is assessed using a grading rubric meticulously designed by educators to evaluate the learner's L2 ability, such as vocabulary and grammatical understanding. 
Such a rubric contains multiple analytic criteria aligned with the specific pedagogical objectives that an educator intends to assess in the question.
This aspect characterizes STE evaluation and distinguishes them from typical GEC tasks, which assess the overall correctness of the grammar.


Evaluation based on the analytic scoring criteria highlights the degree to which the learning objectives are achieved. To this end, some degree of constraint is imposed on the question design and answer choices, limiting the freedom of translation. 
However, if translation variations are observed, all possible expressions are accounted for. 
These restrictions in translation practice, as discussed in \cite{oro21266}, prevent learners from easily avoiding knowledge gaps and direct their attention to L2 aspects that they may find challenging. Therefore, these constraints can be useful in focusing students' attention on specific language abilities.

In addition, the evaluation of translated sentences in educational settings is also different from that of general translations in that the former involves pedagogical objectives such as the acquisition of specific language knowledge.

\subsection{Task formulation}
\label{ssec:Task Formulation}
The purpose of assessing the STE task is to determine how well students' responses achieve the learning objectives defined by the instructors.
To effectively do so, instructors use a carefully constructed scoring rubric.
Each STE question targets several learning objectives and evaluates other fundamental grammatical items (e.g., number, tense, etc.); therefore, a scoring rubric contains multiple independent analytical criteria to evaluate specific items.
These criteria serve as the basis for grading each student's response, with a corresponding analytic score assigned to each grading item (see Table~\ref{tab:Rubric}).

\begin{table*}[ht!]
\centering
\footnotesize
\begin{tabular}{|l|c|l|l|}
\hline
\multicolumn{1}{|c|}{\textbf{Chunk}} & \textbf{Category} & \multicolumn{1}{c|}{\textbf{Correct (2)}} & \multicolumn{1}{c|}{\textbf{Incorrect (0)}} \\ \hline
\multirow{2}{*}{\begin{tabular}[c]{@{}l@{}}私は \\ (I)\end{tabular}} & E & Expressed as "I" & Otherwise \\ \cline{2-4} 
 & O & Word order is "before + Subject" & Not "before + Subject" \\ \hline
\multirow{2}{*}{\begin{tabular}[c]{@{}l@{}}一昨年に \\ (two years ago)\end{tabular}} & E & Expressed as "two years ago" or ... & \begin{tabular}[c]{@{}l@{}}"in the year before last" \\  and otherwise\end{tabular} \\ \cline{2-4} 
 & O & \begin{tabular}[c]{@{}l@{}}Word order is "in Australia \textless{}chunk\textgreater{}" or ...\end{tabular} & Otherwise \\ \hline
\multirow{2}{*}{\begin{tabular}[c]{@{}l@{}}オーストラリアで\\  (in Australia)\end{tabular}} & E & Expressed as "in Australia" & Otherwise \\ \cline{2-4} 
 & O & \begin{tabular}[c]{@{}l@{}}Word order is "\textless{}chunk\textgreater{} two years ago" or ...\end{tabular} & Otherwise \\ \hline
\multirow{3}{*}{\begin{tabular}[c]{@{}l@{}}見るまで \\ (before I saw)\end{tabular}} & E & \begin{tabular}[c]{@{}l@{}}Expressed as "before I saw one," \\ "before I saw some," or ...\end{tabular} & \begin{tabular}[c]{@{}l@{}}The word "it" is used \\ instead of "one" / ...\end{tabular} \\ \cline{2-4} 
 & O & \begin{tabular}[c]{@{}l@{}}The order is\\  "Conjunction + Subject + Verb + Object"\end{tabular} & Otherwise \\ \cline{2-4} 
 & G & The past tense"saw" is used. & "saw" is not used \\ \hline
\multirow{2}{*}{\begin{tabular}[c]{@{}l@{}}コアラを \\ (a koala)\end{tabular}} & E & Expressed as"a koala, " "koalas," "any koalas,"... & Otherwise \\ \cline{2-4} 
 & O & The word immediately follows "seen" & Otherwise \\ \hline
\multirow{3}{*}{\begin{tabular}[c]{@{}l@{}}見た \\ (seen) \end{tabular}} & E & Expressed as "seen" & Otherwise \\ \cline{2-4} 
 & O & \begin{tabular}[c]{@{}l@{}}It is placed immediately after "never," "not," or "n't."\end{tabular} & \begin{tabular}[c]{@{}l@{}} Otherwise\end{tabular} \\ \cline{2-4} 
 & G & \begin{tabular}[c]{@{}l@{}}The past participle form "seen" is used\end{tabular} & \begin{tabular}[c]{@{}l@{}}Otherwise\end{tabular} \\ \hline
\multirow{3}{*}{\begin{tabular}[c]{@{}l@{}}～ことがなかった\\  (had never)\end{tabular}} & E & Expressed as "I had never," "I had not," ... & \begin{tabular}[c]{@{}l@{}}Expressed as "I have \\ never", ... , and others\end{tabular} \\ \cline{2-4} 
 & O & \begin{tabular}[c]{@{}l@{}}The word order is "Subject + Verb"\end{tabular} & \begin{tabular}[c]{@{}l@{}} Otherwise\end{tabular} \\ \cline{2-4} 
 & G & \begin{tabular}[c]{@{}l@{}}The past perfect tense is used\end{tabular} & \begin{tabular}[c]{@{}l@{}}The present perfect or \\ past tense are used\end{tabular} \\ \hline
\end{tabular}

\caption{Examples of analytic criteria excerpted from Q11: ``\textit{I had never seen a koala before I saw one in Australia two years ago.}''  ``Chunk'' denotes a Japanese phrasal unit, often referred to as ``bunsetsu.'' Every chunk invariably includes the category E (Expression), with some incorporating the categories O (Word Order) and G (Grammar). The analytic criterion covers examples of expressions and structural patterns that correspond to each label.} 

\label{tab:Rubric}
\end{table*}

The automatic scoring of analytic criteria was formulated by \citet{Mizumoto2019-xd} as an analytic score prediction task for reading comprehension questions. 
Therefore, this study also considers the analytic score prediction for the automatic scoring of STE. 

\paragraph{Analytic score prediction:}
For a given STE, let $C$ denote the set of analytic criteria.
For the input response text $(w_1,w_2,...,w_n)$, the model outputs an analytic score $s_c \in \{2, 1, 0\}$ for a given analytic criterion $c \in C$, where 2, 1, and 0 represent ``correct,'' ``partially correct,'' and ``incorrect,'' respectively.


\section{Sentence translation exercise (STEs) dataset}
\label{sec:Data}
To implement the automatic STE scoring, we introduce an STE dataset.
This dataset currently comprises 21 Japanese-to-English STE questions with detailed rubrics and annotated student responses.
These questions and the scoring rubrics were created by specialists in the design of English learning materials. The questions were constructed to cover all the major grammar topics in several well-known English textbooks used in Japanese high schools.

Table~\ref{tab:Rubric} shows an example of a rubric, which contains 17 analytic criteria: three for grammar (labeled as ``G''), seven for vocabulary and expression (labeled as ``E''), and seven for word order (labeled as ``O'').
Each analytic criterion is evaluated on a three-point scale: 2 (correct), 1 (partially correct), and 0 (incorrect); the rubric lists the typical expressions for each scale.

Essentially, STEs are designed such that they limit variations in correct responses from the outset.
In practical settings, however, educators may adjust the grading rubric by incorporating variations in correct responses, previously unidentified during the rubric's initial creation, to accurately evaluate the student responses. 
To replicate this process, we initially create the analytic criteria, followed by the collection of student responses as described in the following subsection. 
Subsequently, we refine the rubric by reviewing the collected responses, to preempt any challenges that may arise during the grading procedure. 





In the following sections, we will discuss in detail the methods used to gather responses, as well as the annotation process, and statistically analyze the whole dataset.

\subsection{Collecting student responses}
Ideally, student responses are compiled within classrooms and other practical learning environments. However, the number of responses that can be collected from actual classrooms is often limited, and the collecting process is time-consuming.
Therefore, we constructed our dataset through a combined approach involving high school students and crowdsourcing workers to collect responses for response collection. 
In this approach, we conducted a pilot data collection in which responses were obtained from high school students. 
Then, we analyzed these responses with English education experts and created the criteria for gathering crowdsourcing workers whose English abilities are equivalent to those of the high school students (see Appendix \ref{sec:appendix_recruite_scriteria} for details regarding the recruitment criteria). 
Finally, we hired workers who met the criteria and allowed them to answer the questions, thus collecting a sufficient amount of responses.


To maintain quality, we manually reviewed the collected responses and excluded those that significantly deviated from the expected responses.
As a result, we obtained an average of 167 responses per question.
The following section will present the statistics of the dataset.


\subsection{Annotation:}
\label{ssec:annotation}
As explained in Section \ref{ssec:Task Formulation}, the scoring task for STEs involves grading on a three-class classification for each analytic criterion. 
Annotators are also asked to identify the specific phrase of the response that serves as a grading clue (referred to as {\it justification cues}).
We annotated both types of information in each response.

We hired professional graders to annotate those responses.
As demonstrated in Figure~\ref{fig:sentence_translation_example},
the annotators assigned an analytic score to the responses based on each analytic criterion. 

\paragraph{Justification cue:}
\citet{Mizumoto2019-xd} also annotated specific substrings within responses that contribute to an analytic score. These substrings are called {\it justification cues} because they serve as the rationale for the analytic scores.
We also annotated justification cues in our dataset to enhance the interpretability of analytic scores.
For example, in Figure~\ref{fig:sentence_translation_example}, the phrase ``before I saw'' was annotated as a justification cue and was assigned an analytic score of ``$0$.'' 


\paragraph{Annotation quality:}
To measure the quality of the annotations, we randomly selected 10 out of the 21 questions and asked a different annotator to annotate 20 responses for each question.
We then used Cohen's Kappa coefficient ~\cite{cohen_kappa_1960} to calculate inter-grader agreement for analytic scoring and the F-score to calculate agreement for justification cues.

The scores for all analytic criteria had an overall average Kappa coefficient of 0.74, indicating substantial agreement~\cite{8d20e0b8-89d8-3d65-bcf5-8c19d56ec4ab}. 
Regarding agreement for justification cues, the F-score was 0.92, signifying a high level of agreement among the annotators~\cite{Mizumoto2019-xd, Sato2022-yq}. This suggests that different annotators can consistently identify the same phrase as a justification cue for an analytic score.

\paragraph{Statistics of data:}
Table\ref{tab:Sta} shows the dataset statistics.
We annotated a total of 3,498 responses for 21 questions, including 196 analytic criteria.
For the pilot question, ranging from Q1 to Q7, scoring included 1~(partially correct) whereas the other questions followed a binary scoring of 2~(correct) and 0~(incorrect).
Additionally, the number of instances with a grade of 0 was relatively fewer than those with a grade of 2. 
This distribution was similar to the one observed in the pilot question and others.
Therefore, we conclude that we have successfully gathered crowdsourcing workers whose English ability is equivalent to that of original high school students and that these workers have attempted to answer those questions correctly.

\begin{table}[t]

\centering
\small
 \begin{tabular}{lrrrrr}
        \toprule
        &\#Ans&\#Criteria& 2 & 1 & 0\\
        \midrule
        Q1 & 159 &  9 & 923 & 114 & 235\\
        Q2 & 172 &  8 & 652 &  98 & 454\\
        Q3 &  77 &  8 & 357 &  40 & 142\\
        Q4 &  69 &  9 & 356 &  76 & 120\\
        Q5 & 102 &  9 & 387 & 161 & 268\\
        Q6 &  79 & 12 & 701 &  14 & 154\\
        Q7 &  90 & 10 & 534 &  72 & 204\\
        Q8 &  200 (173) & 6 & 856 &   & 343\\
        Q9 &  200 (169) & 10 & 1324 &   & 676\\
        Q10 &  200 (180) & 9 & 1197 &   & 612\\
        Q11 &  200 (142) & 10 & 1285 &   & 715\\
        Q12 &  200 (135) & 8 & 1175 &   & 425\\
        Q13 &  200 (137) & 7 & 850 &   & 550\\
        Q14 &  150 (97) & 8 & 847 &   & 353\\
        Q15 &  200 (159) & 11 & 1347 &   & 853\\
        Q16 &  200 (144) & 10 & 1565 &  & 435\\
        Q17 &  200 (162) & 11 & 1082 &   & 1118\\
        Q18 &  200 (162) & 9 & 1220 &   & 580\\
        Q19 &  200 (166) & 12 & 1671 &   & 729\\
        Q20 &  200 (149) & 8 & 1064 &  & 536\\
        Q21 &  200 (131) & 12 & 1538 &  & 862\\
        \bottomrule
 \end{tabular}
 \caption{STE dataset statistics. The integers 2, 1, and 0 stand for ``correct,'' ``partially correct,'' and ``incorrect'' labels, respectively. Q8 through Q21 include some identical responses following the distribution of the collected data. We show the number of distinct responses in parentheses.}
 \label{tab:Sta}
\end{table}

\section{Method}
\label{sec:Method}
We employ a BERT~\cite{Devlin2019-yo}-based classification model and the GPT models~\cite{openai2023gpt4} with in-context learning as a baseline for our task formulation.
This section discusses these baseline models in detail.


\subsection{Finetuned BERT model}
\label{ssec:Bert_model}
We employ BERT, which is widely used in various NLP tasks, including SAS, as a baseline for this task.
This model is finetuened for each scoring item in the rubric using the training data.

\paragraph{Architecture:}
First, the response text sequence $\mathbf{w} = (w_{\text{cls}}, w_1, w_2, ..., w_n)$, with a prepended CLS token, is input into BERT, obtaining the intermediate representation $\mathbf{h} = (h_{\text{cls}}, h_1, h_2, ..., h_n)$ as follows:
\begin{equation}
\mathbf{h} = \texttt{BERT}(\mathbf{w})
\end{equation}

In our task, a justification cue that indicates the rationale behind its score is provided for each response.
By utilizing this justification cue to train a model, we expect that the model will grade faithfully according to the rubric.
Therefore, following ~\citet{Mizumoto2019-xd}, we use these justification cues as supervisory signals to train the model's attention layer.
Here, we perform pooling on the BERT-encoded representations using a Bi-LSTM and attention mechanism.
The sequence obtained from $\mathbf{h}$ by excluding $h_{\text{cls}}$ is input into the Bi-LSTM, yielding $\mathbf{h'} = {h'_1, h'_2,...,h'_n}$.
Then we calculate the weighted sum as follows:
\begin{equation}
    \tilde{h_c}=\sum_{i=1}^{n}{\alpha^c_ih'_i},
\end{equation}
where $\alpha^c_i$ is the weight of the $i$-th word relative to the scoring rubric $c$, calculated by the attention mechanism shown in Equation~(\ref{eq:attention}).
\begin{eqnarray}
\label{eq:attention}
t^c_i&=&{h_i}M_cV_c \nonumber \\
\alpha^c_i&=&\frac{\exp(tanh(t^c_i))}{\sum_{k=1}^n{\exp(tanh(t^c_k))}},
\end{eqnarray}
where $M_c\in\mathbb{R}^{D \times D}$ and $V_c\in\mathbb{R}^{D}$ are learnable parameters. Finally, the evaluation value $S_C$ for item $C$ is obtained by the following formula:
\begin{eqnarray}
    p(s_c|\mathbf{w})&=&\texttt{softmax}(W\tilde{h_c} + b) \nonumber \\
    s_c&=&\argmax_{s_c\in{\{0,1,2\}}}\{p(s_c|\mathbf{w})\},
\end{eqnarray}
where $W\in\mathbb{R}^{3\times D}$ and $b\in\mathbb{R}^{3}$ are the learnable parameters.

\paragraph{Training:}
\label{sssec:training}
The analytic scoring model is trained to minimize the negative log-likelihood $(\NLL)$ for each analytic score.
\begin{equation}
    L_{score}=\sum_{c\in{C}}{\NLL(p(s_c|\mathbf{w}),\hat{s_c})}
\end{equation}
where $s_c$ is the label (evaluation value) of the ground truth for scoring rubric $c$.
In addition, as discussed in the Section~\ref{ssec:annotation}, the dataset contains the justification cues $\hat{\alpha^c} =(\hat{\alpha^c_1},\hat{\alpha^c_2},...,\hat{\alpha^c_n})$ for each analytic criterion for the response, where
$\hat{\alpha^c_i} \in [0,1]$ is the indicator of whether the $i$-th token in the response is the justification cue for the score of the analytic criterion $c$.
When the gold justification cue includes $k$ tokens, the sum of $\hat{\alpha^c}$ is $k$. Therefore, as a gold signal for $\alpha^c$, we use $\hat{\alpha^c}$ divided by $k$ during the training process.
Following~\citet{Mizumoto2019-xd}, we use the MSE-based loss function to achieve supervised training of the attentions with justification cues.
\begin{equation}
    L_{att}=\sum_{c\in{C}}\sum_{i=1}^{n}{(\alpha^c_i - \hat{\alpha^c_i})^2}
\end{equation}
Thus, the overall loss $L$ is expressed as:
\begin{equation}
    L=L_{score}+L_{att}.
\end{equation}


\subsection{GPT models with in-context learning}
\label{ssec:method_gpt-3.5}
We evaluate the GPT-3.5 and GPT-4 models in the setting of few-shot in-context learning~\cite{brown2020language}, which significantly minimizes the cost of building a scoring model specific to each grading item as well as the training examples required for finetuning.
Furthermore, the GPT series demonstrates superior performance in tasks such as translation and summarization, among other tasks~\cite{gladkoff2023predicting,helwan2023medical}. 
Therefore, we can expect the proficiency in grammatical knowledge required for automatic grading of STEs.

\begin{figure}[t]
  \centering
  \includegraphics[width=75mm]{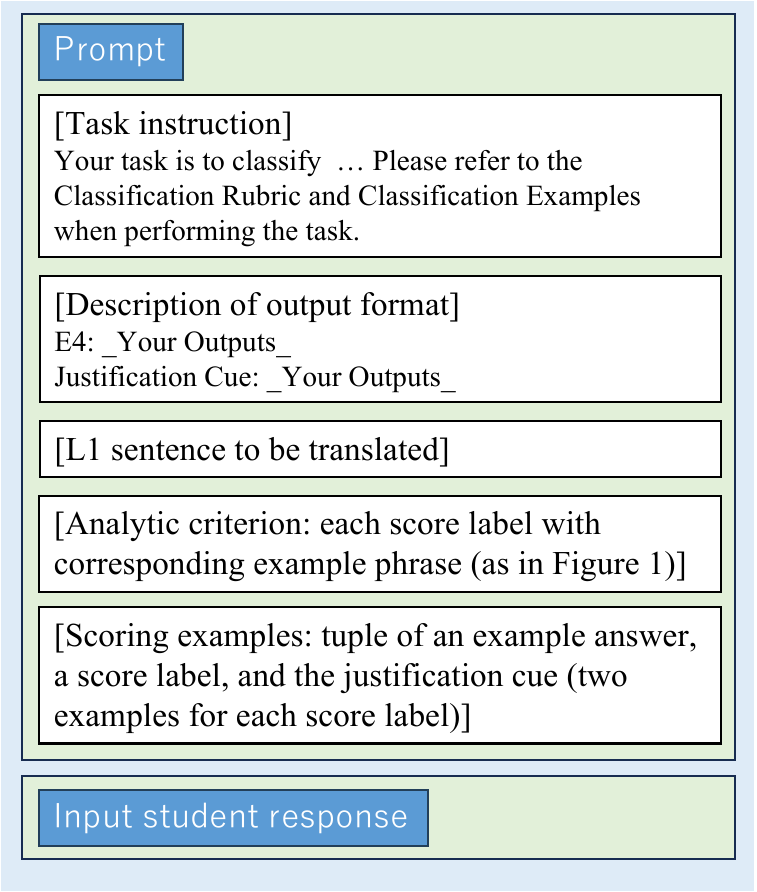}
  \caption{Input for the GPT models}
  \label{fig:prompt_for_gpt}
\end{figure}

Figure~\ref{fig:prompt_for_gpt} shows the input template for the GPT models. 
The input can be segmented into two parts. 
The first part is a {\it prompt} that includes a task instruction, a description of the output format, an L1 sentence for translation, a focused single analytic criterion, and the scoring examples corresponding to that criterion.
The analytic criterion is a (literal) textual representation of a rubric item described in a single row in Table~\ref{tab:Rubric}. 
For each score label, we provide a few-shot examples to illustrate the analytic criterion and its scoring (output examples) for in-context learning.
The second part is a {\it student response}. 
The model leverages these two inputs to generate a score label for the specified criterion and identify the substring of the student response that justifies the evaluation. 
In the GPT models, we treat the grading of each analytic criterion within a prompt as an independent grading task, thus the GPT models output a score for each analytic criterion independently.
More details of the input prompt can be found in Table~\ref{tab:Example_of_prompt_to_gpt} in the appendix.

\begin{table*}[ht!]
\centering
\begin{tabular}{ccccccc}
\toprule
\multicolumn{1}{l}{\multirow{2}{*}{\textbf{\begin{tabular}[c]{@{}l@{}}Category \\ (\#criteria)\end{tabular}}}} & \multicolumn{3}{c}{\textbf{BERT}} & \multicolumn{3}{c}{\textbf{GPT-3.5 (2 shots)}} \\
\cmidrule(lr){2-4}\cmidrule(lr){5-7}
\multicolumn{1}{l}{} & Correct  & Partial. Correct& Incorrect & Correct & Partial. Correct & Incorrect \\ 
\midrule
\textbf{E : (96)} & 0.92\small{± 0.15}& 0.64\small{±0.36}& 0.82\small{±0.24}& 0.83\small{± 0.12}& 0.80\small{±0.23}& 0.62\small{±0.20} \\
\textbf{O : (42)} & 0.95\small{±0.05}& nan & 0.79\small{±0.25}& 0.78\small{±0.11}& nan & 0.52\small{±0.21} \\
\textbf{G : (45)} & 0.94\small{±0.11}& 0.81\small{±0.21}& 0.88\small{±0.13}& 0.81\small{±0.13} & 0.48\small{±0.11}& 0.63\small{±0.25}\\
\midrule
\textbf{All} & 0.93 & 0.68 & 0.83 & 0.81 & 0.73 & 0.59 \\
\bottomrule

\toprule
\multicolumn{1}{l}{\multirow{2}{*}{\textbf{\begin{tabular}[c]{@{}l@{}}Category \\ (\#criteria)\end{tabular}}}} & \multicolumn{3}{c}{\textbf{GPT-3.5 (5 shots)}} & \multicolumn{3}{c}{\textbf{GPT-4 (2 shots)}} \\
\cmidrule(lr){2-4}\cmidrule(lr){5-7}
\multicolumn{1}{l}{} & Correct  & Partial. Correct& Incorrect & Correct & Partial. Correct & Incorrect \\ 
\midrule
\textbf{E : (96)} & 0.84\small{± 0.12}& 0.79\small{±0.23}& 0.65\small{±0.18}& 0.91\small{± 0.09}& 0.80\small{±0.15}& 0.78\small{±0.20} \\
\textbf{O : (42)} & 0.80\small{±0.12}& nan & 0.53\small{±0.21}& 0.87\small{±0.08}& nan & 0.65\small{±0.21} \\
\textbf{G : (45)} & 0.82\small{±0.13}& 0.48\small{±0.11}& 0.64\small{±0.28}& 0.90\small{±0.08} & 0.62\small{±0.37}& 0.77\small{±0.24}\\
\midrule
\textbf{All} & 0.83 & 0.73 & 0.61 & 0.89 & 0.76 & 0.73 \\
\bottomrule

\end{tabular}
\caption{$F_1$ scores and standard deviations of the baseline models for each score label of the analytic criteria categories (E: Expression, O: Word Order, G: Grammar). The analytic criteria for the Word Order category do not include any partially correct expressions; therefore, the corresponding values are represented as ``nan.''}
\label{tab:all_score}
\end{table*}

\section{Experiments}
\label{sec:experiments}
In the experiment, we investigate the feasibility of our task formulation for STEs using the BERT model and the state-of-the-art large language models, GPT-4 and GPT-3.5. 
 We also investigate the impact of the number of in-context examples on the scoring performance.

\subsection{Settings}
In our dataset, the label ``partially correct'' was infrequently used, which transformed the grading of certain criteria into a binary classification task.
Therefore, we used the $F_1$-score to evaluate the performance of the analytic score prediction as it applies to both three-class and binary classification.
We also performed a 5-fold cross-validation by dividing the dataset of each question into a training set, a development set, and an evaluation set following a 3:1:1 ratio. 

We finetuned the BERT model (described in Section~\ref{sec:Method}) for 50 epochs on each training set. 
For each epoch, we calculated $F_1$-score for each analytic criterion and used the parameters that produced the best results on the development set for each analytic criterion, respectively.
Appendix~\ref{sec:appendix_hyper_parameter} provides details regarding these hyperparameters. 
For the GPT models, we randomly selected few-shot examples for each score from the training set.

Some analytic criteria contained extremely few incorrect responses because typical high school students found them too easy. Therefore, to ensure a proper performance evaluation, we used only those criteria that contained 10\% or more incorrect instances.



\subsection{Results}
\label{ssec:results}
Table~\ref{tab:all_score} shows the performance of BERT, GPT-3.5, and GPT-4 on the test set in terms of $F_1$ averages and standard deviations for each category (\textit{Expression, Word Order, Grammar}).

In Section~\ref{ssec:method_gpt-3.5}, we hypothesized that the GPT models would demonstrate excellent performance because STEs evaluate the validity of English sentences within a highly limited grammar and vocabulary scope presented in an analytic criterion.
Surprisingly, however, the BERT model outperformed the GPT models on our dataset.

Nevertheless, both models showed relatively high performance in grading correct responses.
Meanwhile, the GPT models performed notably lower in grading incorrect responses.
Interestingly, however, the GPT models outperformed BERT in grading partially correct responses. 
This may be due to the limited data size for fine-tuning the BERT model for partially correct responses.
We also observed that the standard deviation exceeded 0.10 for nearly all results, indicating a substantial variance in grading performance across different analytic criteria, some of which showed poor results.
The result suggests that the grading of several analytic criteria is challenging for models.

LLMs acquire sufficient knowledge about language, including grammar and vocabulary, through pretraining on massive corpora. However, these results showed that STEs grading remains a challenging task even for a cutting-edge LLM such as GPT-4, when provided with only few-shot examples.
Furthermore, collecting and annotating enough responses to train the STE grading model poses a significant burden in actual educational settings, allowing room for improvement in deploying automatic grading models in actual classrooms.

\subsection{Analysis}

\paragraph{Lower performance for incorrect responses:}
As discussed in Section~\ref{ssec:results}, the models showed notably lower performance in grading incorrect responses than in grading correct responses.
This discrepancy may be due to the difference in the number of variations between correct and incorrect responses. 
As shown in Figure \ref{fig:sentence_translation_example}, the variation of acceptable correct responses is limited; meanwhile, the variation of incorrect responses shows considerable latitude, potentially encompassing any type of response besides the correct ones.
Consequently, although the training data covered the majority of variations in correct responses, they cannot cover all potential incorrect responses.
Additionally, the GPT models significantly struggled in grading such incorrect responses, especially with fewer examples than the BERT models.

\begin{table}[t]
\small
\centering
\begin{tabular}{l}
\toprule
\textbf{Input summary}  \\ 
\midrule
\begin{tabular}[c]{@{}l@{}}Sentence:\\
私は / 一昨年に / オーストラリアで / 見るまで /\\
コアラを / 見た / ことがなかった \\
(I / the year before last / in Australia / before I saw one \\/ a koala / seen / had never)\\\\
Analytic criteria: G1 (Tense) \\
- Past tense "saw" is used as a verb\\\\
Student answer:\\ 
I had never seen a koala before I have seen it \\two years ago in Australia . \\ 
\end{tabular} \\


\toprule

\textbf{GPT output \& (gold data)}  \\ 
\midrule
\begin{tabular}[c]{@{}l@{}}Label: 2\\
(Gold label: 0)\\
Justification cue:  I had never\\
(Gold justification cue: seen)\\
\end{tabular}\\

\bottomrule
\end{tabular}
\caption{Example of a prediction error made by GPT-3.5. }
\label{tab:error}
\end{table}

\begin{figure}[t]
  \centering
  \includegraphics[width=75mm]{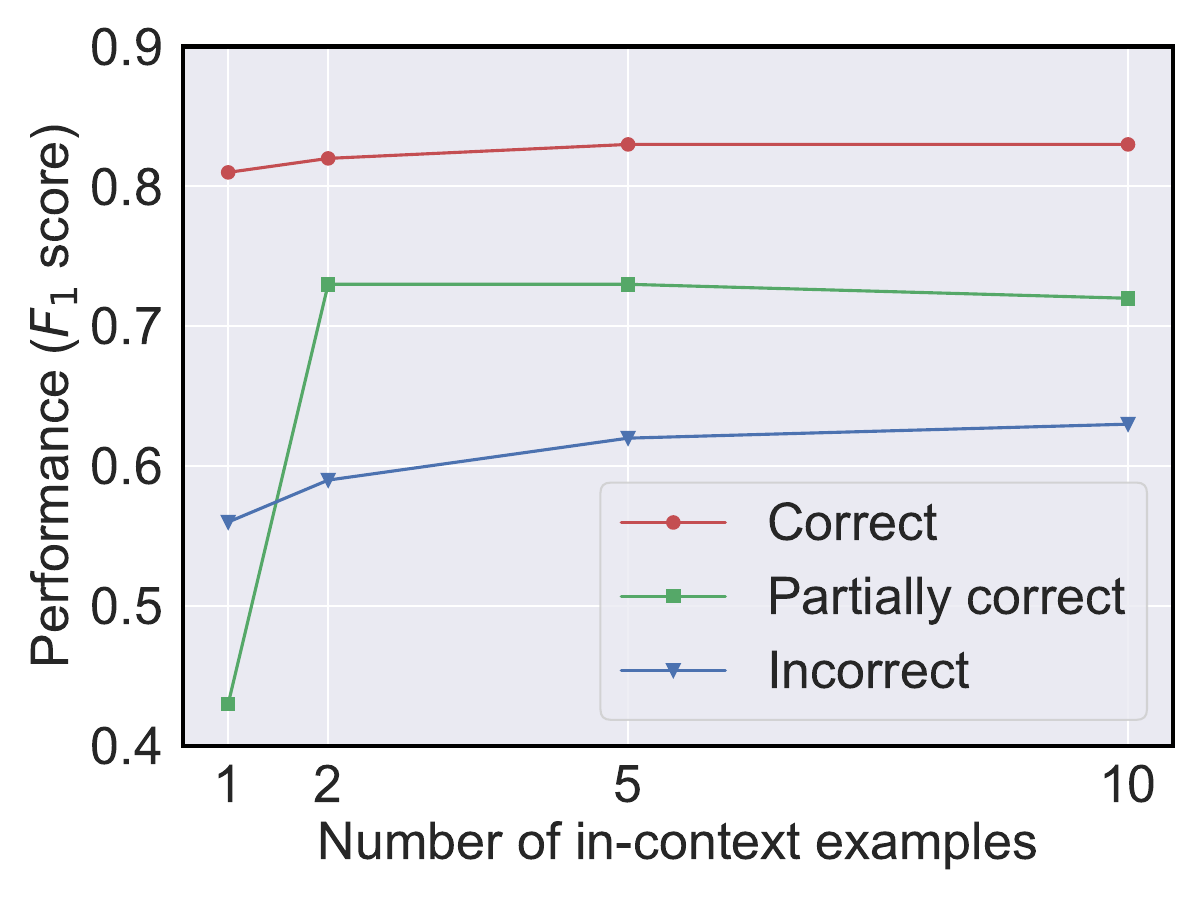}
  \caption{The performance of the GPT-3.5 model when changing the number of in-context examples. The x-axis represents the number of in-context examples. The y-axis represents the averaged $F_1$-score among all analytic criteria.}
  \label{fig:Nshot-curve.pdf}
\end{figure}

\paragraph{Grading error example of GPT-3.5:}
Table~\ref{tab:error} shows a grading error made by GPT-3.5, in which the model significantly failed to recognize an incorrect response.
Such grading errors constitute the majority of inaccurate predictions by GPT-3.5.
We hypothesized that these inaccuracies are due to the specialized prompt and response format of STEs, including scores, detailed rubrics, and justification cues.
Hence, during pretraining, GPTs are not exposed to such a task, despite the extensive corpora collected from the Web.
Utilizing GPT-3.5 for few-shot in-context learning is expected to be more suitable for classroom applications than fine-tuning the model with a substantial amount of training data. 
However, our observations suggest that this application of GPT-3.5 is inadequate for grading STEs. 
\paragraph{The impact of the number of in-context examples:}
To investigate the appropriate number of in-context examples, we evaluate performance by varying the number of examples provided in the prompt.
Figure \ref{fig:Nshot-curve.pdf} illustrates the $F_1$-score of GPT-3.5 for each label as the number of in-context examples is varied between one, two, five, and 10.  
From the result, we can clearly see that the grading performance hardly changed even when the number of in-context examples was increased to more than two.

As a reason for this, in grading for correct responses, it is considered that our task design inherently results in a very limited number of patterns corresponding to correct expressions. Therefore, increasing the number of instruction samples may not significantly influence the accuracy for correct responses.

In the grading of incorrect responses, a considerable number of instances are labeled as incorrect due to the absence of expressions equivalent to the correct answers.
In such cases, \textit{justification cue} string is not given in the instruction for GPTs and this makes it challenging to grasp scoring clues from the provided instruction examples, likely hindering the effective learning of appropriate grading and consequently impeding performance improvement.

\section{Related work}

Grammar Error Correction (GEC) and Short Answer Scoring (SAS) are the two major research areas in the automatic evaluation of descriptive English responses. 
We position this study between these two research domains.

\subsection{Grammar Error Correction (GEC)}
The most famous GEC system is Grammarly,\footnote{\url{https://www.grammarly.com}} a writing assistant tool that also plays an important role in English learning~\cite{Ranalli2021-sl,Koltovskaia2020-je}.
In a more educational context, ~\citet{nagata-2019-toward} proposed the task of feedback generation in GEC with a focus on effective ESL (English as a Second Language)  learning.
Some studies have also focused on methods to generate feedback for grammatical errors in sentences written by learners~\cite{hanawa-etal-2021-exploring,coyne-2023-template,lai-chang-2019-tellmewhy}.
Regarding the use of LLMs in GEC, \citet{fang2023chatgpt} reported that GPT-3.5 shows excellent GEC abilities.


All these previous studies have focused primarily on identifying grammatical errors present in freely-composed text.
However, within real-world educational contexts that require the measurement of student progress in language learning, educators must direct their attention to the assessment of not only overarching grammatical constructs but also a precise understanding of certain grammatical or vocabulary items within specific units of English textbooks.
Such a methodology would determine students' comprehension and areas of unfamiliarity more accurately. 
Therefore, we adopted this practical approach by developing STEs specifically designed to evaluate students' understanding of various grammatical topics. 




\subsection{Short Answer Scoring (SAS)}
We have formally defined our STE grading task within the established framework of the automated SAS task. 
However, these two tasks fundamentally differ in terms of their intended objectives and the descriptive content to be evaluated.
Several SAS studies have primarily examined closed-domain questions that require knowledge and understanding in specific areas, such as science or reading comprehension~\cite{Mizumoto2019-xd,Burrows2015-vs,Galhardi2018-xq}, and a typical SAS framework does not directly consider grammatical errors and word usage errors in responses. 
In this study, we created detailed and stringent analytic criteria for measuring learners' English proficiency, focusing on the grammatical aspects addressed in the questions.

\paragraph{Dataset:}
The dataset we created for the STE task followed the format of the RIKEN SAS dataset, which contains questions on Japanese reading comprehension questions~\cite{Mizumoto2019-xd, Funayama2023-pr}. 
Other SAS datasets include \textsc{BEETLE} ~\cite{noauthor_undated-ay}, \textsc{ASAP-SAS},\footnote{\url{https://www.kaggle.com/c/asap-sas/}} \textsc{Powergrading}, and the \textsc{SAF dataset}~\cite{filighera-etal-2022-answer}, which focus on science or reading comprehension. 
Our dataset is the first STE dataset to concentrate on grading grammar and vocabulary use.


\section{Conclusion}
This study introduced a novel task focusing on the automatic grading of Sentence Translation Exercises (STEs) for educational purposes.
We formalized STEs as a task of grading each analytic criterion predetermined by teachers' intentions and constructed a dataset to implement the task.
This first-of-its-kind dataset emulates and reflects the practical form of L2 learning in the responses of learners. We also used finetuned BERT and GPTs with few-shot in-context learning to establish a baseline and demonstrate the feasibility of the formulated framework.

In our experiment, although the GPT models showed substantial performance in various NLP tasks, they remained inferior to the BERT model, suggesting that our newly defined task continues to be challenging even for the state-of-the-art LLMs, therefore necessitating further exploration.

With regard to future direction, we are contemplating the integration of technologies such as GEC and machine translation within our model. We aim to build cross-questions strategies to automatically identify expressions that diverge from a provided rubric while preserving the text's fundamental meaning using a combination of these technologies.
For this purpose, our plan involves further subdividing the STE grading task and leveraging LLMs to address each minimized task such as correcting grammatical errors, assessing the consistency of meaning with L1, and identifying expressions aligned with the learning objectives in each exercise.
This approach also aims to investigate tasks where LLMs may not excel in STE scoring and enhance their overall performance. 
Additionally, in an educational context, we also consider generating more comprehensive feedback comments on the scoring results, extending beyond the estimation of justification cues.


\section*{Limitations}
This section discusses the limitations of our study from the perspectives of dataset creation and experimentation.

\paragraph{Dataset creation:}
We created the first STE dataset in this literature, which includes responses with scores and detailed rubrics.
However, our dataset was limited to Japanese-English translation, while STEs can be applied to any language pair.

Furthermore, we conducted crowd-sourcing to gather responses for our dataset, which may differ from student responses in actual classroom settings, despite a carefully controlled crowdsourcing process as described in Section~\ref{ssec:annotation}.
Therefore, the performance of the models when deployed in real education settings, such as English study in school, remains uncertain.

\paragraph{Experimentation:}
We conducted experiments using only the BERT-based model and the GPT models. 
Therefore, the performance of other LLMs, such as  LLaMA, remains unclear, and the effectiveness of fine-tuning these LLMs using parameter-efficient methods such as LoRA~\cite{Hu2021-gw} is also unexplored.



\section*{Ethics statement}
\paragraph{Gathering crowdsourcing workers:}
To collect responses, we recruited crowdsourcing workers and paid them 18 yen for each question they answered.
In our trial, it took them an average of 1 minute to answer a question; therefore, we estimated the workers' pay at around $1,080$ yen per hour, which is nearly equivalent to Japan's minimum wage of $1,004$ yen per hour in 2023.

\paragraph{Hiring annotators:}
To annotate the dataset, we employed professional English educators through a company that conducts trial annotation and calculated the annotating costs at $150$ yen per response, in agreement with the annotators.
We followed the company's wage proposal.

\section*{Acknowledgements}
The authors are grateful to Diana Galvan-Sosa for her valuable and insightful discussions.
Additionally, we extend our thanks to Shitennoji High School and Junior High School and Zoshindo Juken Kenkyu-sha for their cooperation in preparing the dataset.
This work was supported by JSPS KAKENHI Grant Number 22H00524, JST SPRING, Grant Number JPMJSP2114.


\newpage

\appendix
\section{Recruitment criteria}
\label{sec:appendix_recruite_scriteria}
The recruitment criteria for selecting workers are as follows: (1) TOFEL iBT: 55-70, (2) TOEIC L\&R: 550-750, and (3) The National Center Test: 140 points or higher~\footnote{The Center Test is a standardized test included in the entrance examination of almost all universities in Japan.}. In addition, we also conducted a pretest on candidates for crowdsourcing workers, which consisted of 10 easy STE questions, and we only hired those who answered all of them correctly. 

\section{Prompt example for the GPT models}
Table~\ref{tab:Example_of_prompt_to_gpt} shows an example of a prompt in Q11 used for the GPT models.
\begin{table*}[ht!]

\begin{tabular}{l}
\toprule

\textbf{PROMPT(SYSTEM)}                                                                                                                                                                                                                                                                                                                                                                                                                                                                                                                                                                                                                                               \\ \midrule
\textit{\begin{tabular}[c]{@{}l@{}}Your task is to classify the labels corresponding to the analytic criterion from the input response. \\Please refer to the Classification Rubric and Classification Examples when performing the task.\\\\\_Your Outputs\_\\E4: \_Your Outputs\_\\Justification Cue: \_Your Outputs\_\\\\\_ Question\_\\"私は一昨年にオーストラリアで見るまでコアラを見たことがなかった。\\<I / the year before last / in Australia / before I saw one / a koala / seen / had never> "\\\\ \_Analytic criterion\_\\ E4:Tense of expressions corresponding to "見るまで"\\\\  E4: 2 -Express "見るまで" as "before I saw one(s)" , "before I saw some", "before I saw them"\\ E4: 0-Using "it" instead of "one(s)". Otherwise.\\ \\\_Classification Examples\_\\ Ans：I have not seen koalas before I saw them in Australia 2 years ago .\\ E4： 2\\ justification cue：before I saw them\\\\Ans: I had never seen koalas before I saw ones in Australia two years ago .\\ E4: 2\\justification cue: before I saw ones\\\\Ans: I never see koala before I saw that at Australia last year .\\E4: 0\\justification cue: before I saw that\\\\Ans: I had never seen a koala until I saw it in Australia in the year before last . \\E4: 0\\justification cue: until I saw it                                           \end{tabular}}\\ 
\toprule

\textbf{Input student response}  \\ 
\midrule
\begin{tabular}[c]{@{}l@{}}I had never seen a koala before I saw one in Australia the year before last.\\
\end{tabular}\\

\bottomrule
\end{tabular}

\caption{An example of a prompt for grading an analytic criterion for the phrase ``見るまで'' (before I saw). This prompt contains five parts; task instruction, description of the output format, Question (L1 sentence for translation), the analytic criterion, and a few-shot examples. The task instruction, located at the beginning of the prompt, explains the automatic scoring of STEs. The output format description follows the section labeled \textit{\_Your Outputs\_} in the prompt. The Analytic criterion provides representative examples of expressions that are deemed appropriate or inappropriate for the phrase ``見るまで'' (before I saw). We provide two examples for each label in the few-shot examples and inserted descriptions in '< >' for clarification, but these are not included in the actual prompt.}
\label{tab:Example_of_prompt_to_gpt}
\end{table*}
We input the data into GPTs for each analytic criterion independently. We also input in-context examples for each label.

\section{Implementation and hyperparameter}
\label{sec:appendix_hyper_parameter}
We implemented our BERT model\footnote{We used a pretrained model from \url{https://huggingface.co/bert-base-uncased}} using the Hugging Face library~\cite{wolf-etal-2020-transformers}. During the fine-tuning, we used Adam~\cite{Kingma2014-kp} as the optimizer and set the learning rate to 0.001. The dimension of the hidden state in the Bi-LSTM was set to 128. We also used a batch size of 10, as our dataset contained a relatively small amount of training data.

\end{document}